\documentclass[10pt,conference]{IEEEtran}

\usepackage{cite}

\ifCLASSINFOpdf
   \usepackage[pdftex]{graphicx}
   \graphicspath{{figs/}}
   \DeclareGraphicsExtensions{.pdf,.jpeg,.png}
\else
   \usepackage[dvips]{graphicx}
   \graphicspath{{../figs/}}
   \DeclareGraphicsExtensions{.eps}
\fi
\usepackage[cmex10]{amsmath}
\usepackage{amsthm, amssymb}

\usepackage{algorithmic}

\usepackage{array}

\ifCLASSOPTIONcompsoc
  \usepackage[caption=false,font=normalsize,labelfont=sf,textfont=sf]{subfig}
\else
  \usepackage[caption=false,font=footnotesize]{subfig}
\fi
\usepackage{url}

\usepackage{booktabs} 
\usepackage{multirow} 

\usepackage{tikz}
\usetikzlibrary{shapes.geometric, positioning, arrows.meta}

\newif\iffinal
\finaltrue
\newcommand{\cmtid}{208}

\iffinal
\else
\usepackage[switch]{lineno}
\fi

\newif\ifreview
\reviewfalse

\ifreview
\usepackage[normalem]{ulem}

\providecommand{\Add}[1]{{\color{blue} \uline{#1}}}
\providecommand{\Del}[1]{{\color{red} \sout{#1}}}
\else
\providecommand{\Add}[1]{#1}
\providecommand{\Del}[1]{\empty}
\fi
\providecommand{\Rep}[2]{{\Del{#1}\Add{#2}}}

\begin{document}
%
\thispagestyle{empty}
\onecolumn
\linespread{1.2}\selectfont{}
{\noindent\Huge IEEE Copyright Notice}\\[1pt]

{\noindent\large Copyright (c) 2025 IEEE

\noindent Personal use of this material is permitted. Permission from IEEE must be obtained for all other uses, in any current or future media, including reprinting/republishing this material for advertising or promotional purposes, creating new collective works, for resale or redistribution to servers or lists, or reuse of any copyrighted component of this work in other works.}\\[1em]

{\noindent\Large Accepted to be published in: 2025 38th SIBGRAPI Conference on Graphics, Patterns and Images (SIBGRAPI'25), September 30 -- October 3, 2025.}\\[1in]

{\noindent\large Cite as:}\\[1pt]

{\setlength{\fboxrule}{1pt}
 \fbox{\parbox{0.65\textwidth}{S. F. Santos, T. A. Almeida, and J. Almeida, ``E-MLNet: Enhanced Mutual Learning for Universal Domain Adaptation with Sample-Specific Weighting'' in \emph{2025 38th SIBGRAPI Conference on Graphics, Patterns and Images (SIBGRAPI)}, Salvador, BA, Brazil, 2025, pp. 1--6}}}\\[1in] 
 
{\noindent\large BibTeX:}\\[1pt]

{\setlength{\fboxrule}{1pt}
 \fbox{\parbox{0.95\textwidth}{
 @InProceedings\{SIBGRAPI\_2025\_Santos,
 
 \begin{tabular}{lll}
  & author    & = \{S. F. \{Santos\} and 
                    T. A. \{Almeida\} and
                    J. \{Almeida\}\},\\
			   
  & title     & = \{E-MLNet: Enhanced Mutual Learning for Universal Domain Adaptation with \\
  &           & \ \ \ \ Sample-Specific Weighting\}, \\
			   
  & pages     & = \{1--6\},\\
  
  & booktitle & = \{2025 38th SIBGRAPI Conference on Graphics, Patterns and Images (SIBGRAPI)\},\\
  
  & address   & = \{Salvador, BA, Brazil\},\\
  
  & month     & = \{September 30 -- October 3\},\\
  
  & year      & = \{2025\},\\
  
  & publisher & = \{\{IEEE\}\},\\
  
  \end{tabular}
  
\}
 }}}

\twocolumn
\linespread{1}\selectfont{}
\clearpage

%
\title{E-MLNet: Enhanced Mutual Learning for Universal Domain Adaptation with Sample-Specific Weighting}


\iffinal

\author{
\IEEEauthorblockN{
Samuel Felipe dos Santos,
Tiago Agostinho de Almeida, and
Jurandy Almeida
}\\
\IEEEauthorblockA{
Department of Computing, Federal University of S\~{a}o Carlos (UFSCar), Sorocaba, SP -- Brazil\\
Emails: {\small\texttt{\{samuel.felipe, talmeida, jurandy.almeida\}@ufscar.br}}
}
}


%

\fi

\maketitle

\begin{abstract}
Universal Domain Adaptation (UniDA) seeks to transfer knowledge from a labeled source to an unlabeled target domain without assuming any relationship between their label sets, requiring models to classify known samples while rejecting unknown ones. Advanced methods like Mutual Learning Network~(MLNet) use a bank of one-vs-all classifiers adapted via Open-set Entropy Minimization~(OEM). However, this strategy treats all classifiers equally, diluting the learning signal. We propose the Enhanced Mutual Learning Network~(E-MLNet), which integrates a dynamic weighting strategy to OEM. By leveraging the closed-set classifier's predictions, E-MLNet focuses adaptation on the most relevant class boundaries for each target sample, sharpening the distinction between known and unknown classes. We conduct extensive experiments on four challenging benchmarks: Office-31, Office-Home, VisDA-2017, and ImageCLEF. The results demonstrate that E-MLNet achieves the highest average H-scores on VisDA and ImageCLEF and exhibits superior robustness over its predecessor. E-MLNet outperforms the strong MLNet baseline in the majority of individual adaptation tasks---22 out of 31 in the challenging Open-Partial DA setting and 19 out of 31 in the Open-Set DA setting---confirming the benefits of our focused adaptation strategy.
\end{abstract}

\IEEEpeerreviewmaketitle

\section{Introduction}
\label{sec:introduction}

Deep learning models have achieved remarkable success, yet their performance often relies on supervised learning, which assumes that training and testing data are drawn from the same distribution and share identical classes. These ideal conditions are rare in real-world applications, which frequently face challenges like costly data annotation and distribution shifts caused by dynamic environments where novel classes can emerge~\cite{silva2023sibgrapi,silva2025prl}.

\begin{figure}[!t]
    \centering
    \resizebox{0.9\columnwidth}{!}{
    \begin{tikzpicture}[scale=0.65]
    
    \tikzstyle{source_boundary} = [draw=black, thick, circle, minimum size=3cm]
    \tikzstyle{target_boundary} = [draw=orange, thick, densely dashed, ellipse, minimum width=3.1cm, minimum height=3.1cm]
    
    \tikzstyle{class_shape} = [draw=blue!50!black, fill=blue!60, fill opacity=0.8]
    \tikzstyle{class_circle}   = [class_shape, circle, scale=1.75]
    \tikzstyle{class_square}   = [class_shape, regular polygon, regular polygon sides=4, scale=1.5]
    \tikzstyle{class_triangle} = [class_shape, regular polygon, regular polygon sides=3]
    \tikzstyle{class_star}     = [class_shape, star, star points=5, star point ratio=0.5, fill=blue!60, draw=red!50!black, scale=2]
    \tikzstyle{class_pentagon} = [class_shape, regular polygon, regular polygon sides=5, fill=blue!60, draw=green!50!black, scale=1.5]
    
    \tikzstyle{title_style} = [font=\Large\bfseries]
    \tikzstyle{main_title}  = [fill=black, text=white, font=\Large\bfseries, rectangle, rounded corners]
    \tikzstyle{legend_style}= [font=\sffamily]
    \tikzstyle{annot_style} = [font=\sffamily, align=center]

    \node[title_style] at (0, 3) {CDA};
    \node[source_boundary] (cs_source) at (0,0) {};
    \node[target_boundary] (cs_target) at (0,0) {};
    \node[class_circle]   at (-1, 0) {};
    \node[class_square]   at (0.9, 0.6) {};
    \node[class_triangle] at (0.9, -0.6) {};

    \node[title_style] at (7, 3) {PDA};
    \node[source_boundary] (p_source) at (7,0) {};
    \node[target_boundary, minimum width=1.5cm, minimum height=2.5cm] (p_target) at (7.9, 0) {};
    \node[class_circle] at (6, 0) {}; 
    \node[class_square]   at (7.9, 0.6) {};
    \node[class_triangle] at (7.9, -0.6) {};

    \node[title_style] at (0, -3) {ODA};
    \node[source_boundary] (os_source) at (-0.5, -6) {};
    \node[target_boundary, minimum width=4cm, minimum height=3.3cm] (os_target) at (0.2, -6) {};
    \node[class_circle]   at (-1.5, -6) {};
    \node[class_square]   at (0.4, -5.4) {};
    \node[class_triangle] at (0.4, -6.6) {};    
    \node[class_star]     at (2.4, -5.4) {}; 
    \node[class_pentagon] at (2.4, -6.6) {}; 
    \draw[blue!70!black, thick] (2, -7.1) -- (2.8, -7.1); 
    \draw[-{Stealth[length=3mm]}, blue!70!black, thick] (2.4, -7.1) -- (2.4, -7.6); 
    \node[annot_style] at (2.4, -8.1) {Unknown \\ Categories};

    \node[title_style] at (7, -3) {OPDA};
    \node[source_boundary] (op_source) at (6.5, -6) {};
    \node[target_boundary, minimum size=3cm] (op_target) at (8.4, -6) {};
    \node[class_circle]   at (5.5, -6) {};   
    \node[class_square]   at (7.4, -5.4) {};   
    \node[class_triangle] at (7.4, -6.6) {}; 
    \node[class_star]     at (9.4, -5.4) {};   
    \node[class_pentagon] at (9.4, -6.6) {};   

    \draw[gray!50] (-3, -2.5) -- (10.8, -2.5);
    \draw[gray!50] (3.9, 3.5)   -- (3.9, -9);
    \node[main_title] at (3.9, -2.5) {UniDA};

    \node[source_boundary, scale=0.3] at (2, -9.5) {};
    \node[legend_style, right] at (3.0, -9.5) {Source};
    
    \node[target_boundary, scale=0.3] at (6.5, -9.5) {};
    \node[legend_style, right] at (7.5, -9.5) {Target};
    
    \end{tikzpicture}
    }
    \caption{Different settings for Unsupervised Domain Adaptation~(UDA) based on the relationship between the class sets of the source (solid line) and target (dashed line) domains: Closed-set (CDA), Partial (PDA), Open-Set (ODA), and Open-Partial (OPDA). Universal Domain Adaptation (UniDA) generalizes these settings by assuming this relationship is unknown a priori~\cite{saito2020universal}.}
    \label{fig:da_settings}    
\end{figure}
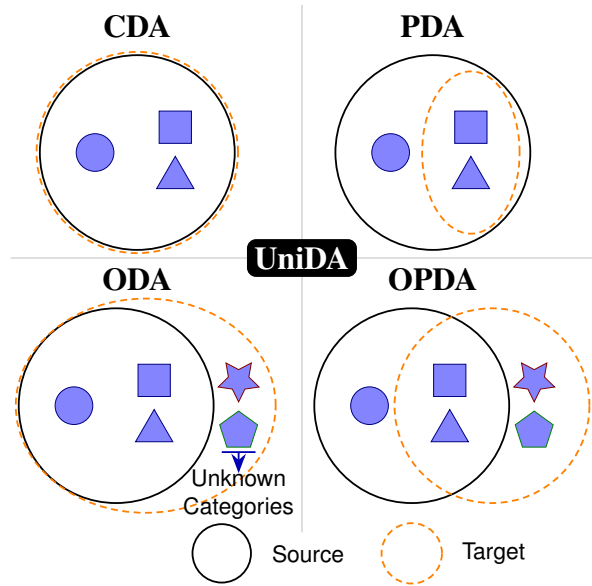

Unsupervised Domain Adaptation~(UDA) offers a strategy to overcome these issues by transferring knowledge from a label-rich source domain to an unlabeled target domain~\cite{silva2021sibgrapi}. However, traditional UDA often operates under the restrictive Closed-Set Domain Adaptation~(CDA) assumption, where both domains share the same set of classes~\cite{saito2020universal}. As illustrated in Figure~\ref{fig:da_settings}, real-world scenarios are more complex, including Partial Domain Adaptation~(PDA), where the target label set is a subset of the source's; Open-Set Domain Adaptation~(ODA), where the target contains novel classes; and a mixture of both, called Open-Partial Domain Adaptation~(OPDA).


Universal Domain Adaptation~(UniDA)~\cite{you2019universal} addresses this challenge by making no assumptions about the label set relationship. A UniDA model must correctly classify target samples from shared classes while simultaneously identifying samples from novel target classes as ``unknown''. Recent methods like OVANet~\cite{saito2021ovanet} and MLNet~\cite{lu2024mlnet} tackle this by training a bank of one-vs-all (open-set) classifiers. These models learn class-specific decision boundaries and adapt them to the target domain using Open-set Entropy Minimization~(OEM). MLNet further enhances this with Neighborhood Invariance Learning~(NIL), Cross-domain Manifold Mixup~(CMM), and a Consistency Constraint~(CC)~\cite{lu2024mlnet}.

In both OVANet and MLNet, OEM is performed by averaging the entropy from all open-set classifiers, implicitly assuming that each one is equally important for any given sample.
Often, only a small subset of class boundaries is truly relevant for making an accurate in-lier/out-lier decision. By treating all classifiers equally, the learning signal is diluted by the vast number of irrelevant classifiers, especially as the number of source classes grows. 
This can lead to suboptimal adaptation and less precise decision boundaries.

Motivated by this observation, we propose the Enhanced Mutual Learning Network~(E-MLNet) for UniDA, which refines the adaptation process of MLNet by integrating a dynamic, sample-specific weighting strategy for OEM.
Inspired by~\cite{yu2024selfcalibrated}, E-MLNet leverages the predictions from the closed-set classifier to weight the contribution of each open-set classifier's entropy term.
This \Rep{creates an attention mechanism that}{dynamic weighting strategy} encourages the model to focus on the most informative open-set classifiers for each target sample, thereby sharpening the learning signal and improving adaptation.

The main contributions of this paper are:
\begin{itemize}
\item We introduce E-MLNet, which enhances MLNet with a \Add{dynamic} weighting strategy to enable the OEM to focus adaptation on the most relevant class boundaries for each target sample, \Rep{leading to more discriminative feature representations and precise unknown-class identification}{improving the model’s robustness and consistency across a wide variety of adaptation tasks.}
\item We conduct extensive experiments on four challenging and publicly available benchmarks for domain adaptation. Our results demonstrate that E-MLNet performs better than MLNet in most of the evaluated scenarios.
\end{itemize}

\section{Related Work}
\label{sec:related}
Universal Domain Adaptation~(UniDA) addresses the challenge of transferring knowledge from a labeled source domain to an unlabeled target domain under the assumption that the relationship between their label sets is unknown. Consequently, the source and target domains may share a set of common classes, while each may also possess private classes. An effective UniDA model must not only correctly classify target samples belonging to the common classes but also reject samples from private target classes as ``unknown''. A diverse range of strategies has been explored to tackle this problem.

One prevalent approach relies on adversarial training with a domain discriminator to learn domain-invariant features. The Universal Adaptation Network (UAN)~\cite{you2019universal} pioneered this by employing entropy to weight samples and quantify their ambiguity. \Rep{This concept was extended by CMU~\cite{fu2020learning} and I-UAN~\cite{yin2021pseudo}, which introduced}{CMU~\cite{fu2020learning} extended this idea by using} more refined techniques for uncertainty measurement and pseudo-labeling. \Del{More recently, SNAIL~\cite{han2023snail} proposed a semi-separated adversarial learning framework to further improve domain alignment.}

Another significant research direction focuses on exploiting the geometric structure of the feature space. Several methods leverage clustering techniques; for instance, Domain Consensus Clustering~(DCC)~\cite{li2021domain} learns robust class representations through consensus. \Rep{Other approaches, such as PCL~\cite{shan2023prediction} and SPA~\cite{kundu2022subsidiary}}{Others, like CPR~\cite{hur2023learning}}, learn explicit prototypes for common classes and \Del{align target features accordingly. CPR~\cite{hur2023learning} extends this by also learning} reciprocal points to explicitly model the ``unknown'' feature space, thereby pushing private target samples away from the known class prototypes.

Self-supervised learning has also emerged as a promising tool for UniDA. DANCE~\cite{saito2020universal} employs self-supervision to cluster target samples while employing entropy minimization to align source and target features and reject unknown instances. Similarly, ROS~\cite{bucci2020effectiveness} explored image rotation as a pretext task to help discriminate between known and unknown samples. Inspired by SwAV~\cite{caron2020unsupervised}, UniOT~\cite{chang2022unified} formulates UniDA as a unified optimal transport problem, seeking to learn an optimal mapping between target samples and source prototypes.

Differently, some methods integrate multiple strategies, designing sophisticated, multi-component loss functions to address the various facets of the UniDA problem simultaneously. 
\Del{For example,} TNT~\cite{chen2022evidential} evaluates sample-level uncertainty with a mutual nearest neighbors contrastive loss.
GATE~\cite{chen2022geometric} employs geometric adversarial learning for global distribution calibration and subgraph-level contrastive learning for local region aggregation.
\Del{MATHS~\cite{chen2022mutual} relies on contrastive alignment of mutual nearest neighbors and incremental pseudo-classifier discrimination guided by hybrid prototype completion.}
\Del{More recently,} NCAL~\cite{su2023neighborhood} employs credibility-weighted conditional adversarial learning to obtain class-invariant features.

Our work belongs to a promising family of methods based on one-vs-all (open-set) classifiers. OVANet~\cite{saito2021ovanet} first introduced this strategy by training a separate binary classifier for each known class, enabling it to learn an adaptive threshold for rejecting unknowns. Building on this, MLNet~\cite{lu2024mlnet} improves upon OVANet with three key components: Neighborhood Invariance Learning~(NIL) for a more robust feature representation, Cross-domain Manifold Mixup~(CMM) to synthesize features that simulate unknown samples, and a Consistency Constraint~(CC) to foster mutual learning between classifiers.

\section{Our Approach}
\label{sec:approach}
\begin{figure*}[!htb]
    \centering
    \begin{minipage}{0.7\textwidth}
    \centering
    \includegraphics[width=0.9\textwidth]{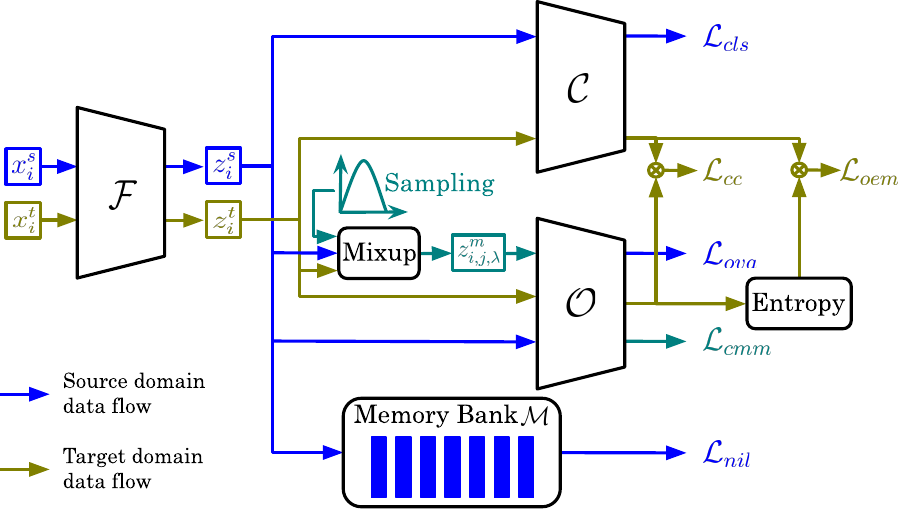}\\
    (a) E-MLNet
    \end{minipage}
    \qquad
    \begin{minipage}{0.25\textwidth} 
    \centering
    \includegraphics[width=0.9\textwidth]{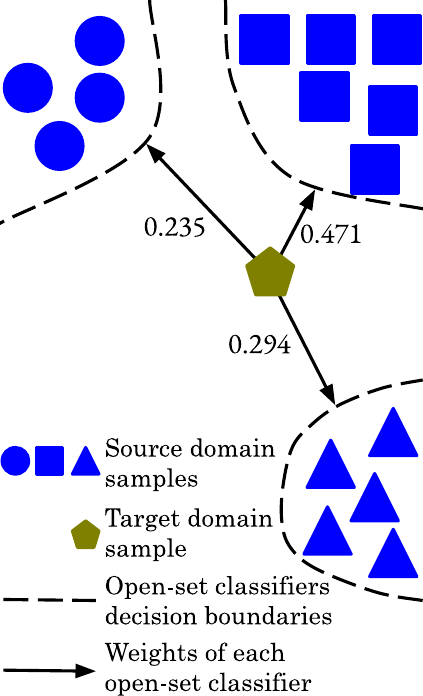}\\
    (b) Feature space
    \end{minipage}

    \caption{Overview of E-MLNet. (a) Schematic of our proposed E-MLNet, highlighting the data flow for the source and target domains and the application of loss functions. (b) Illustration of the feature space with one-vs-all decision boundaries. For a given target sample, its proximity to different class boundaries varies, making some classifiers more relevant than others. Following~\cite{yu2024selfcalibrated}, E-MLNet leverages the closed-set classifier's predictions to weight the importance of each open-set classifier during adaptation.}
    \label{fig:e-mlnet}    
\end{figure*}

This section begins by formally defining the UniDA problem. We then review the architectures and learning objectives of OVANet and MLNet, which serve as the foundation for our method. Finally, we detail our enhancement: a weighting strategy for OEM that improves upon the MLNet framework. Figure~\ref{fig:e-mlnet} shows an overview of our proposed approach.

\subsection{Problem Formulation}
We are given a labeled source domain $D_s = \{(x_i^s, y_i^s)\}_{i=1}^{N_s}$ and an unlabeled target domain $D_t = \{x_j^t\}_{j=1}^{N_t}$. Let $L_s$ and $L_t$ denote the label sets of the source and target domains, respectively. In the UniDA setting, the relationship between $L_s$ and $L_t$ is unknown. The goal is to train a model that can classify target samples $x^t_j$ into one of the known classes present in the source domain ($L_s$) if $x^t_j \in L_s \cap L_t$, or classify it as ``unknown'' if $x^t_j \in L_t \setminus L_s$. For simplicity, we denote $K = |L_s|$ as the number of classes in $L_s$.

\subsection{Preliminaries: From OVANet to MLNet}

Our method builds directly upon the MLNet framework, which itself extends OVANet. The core architecture comprises three components: a feature extractor $\mathcal{F}$, a closed-set classifier $\mathcal{C}$, and a bank of open-set classifiers $\mathcal{O}$. The classifier $\mathcal{C}$ is a standard multi-class classifier trained on the $K$ source classes. The bank of open-set classifiers $\mathcal{O} = \{\mathcal{O}^k\}_{k=1}^K$ consists of $K$ one-vs-all (OVA) binary classifiers. Each classifier $\mathcal{O}^k$ corresponds to a source class $k$ and is trained to distinguish samples of that class (in-liers) from all other classes (out-liers).

\noindent\textbf{OVANet Framework.} OVANet~\cite{saito2021ovanet} trains its components using two primary constraints on source data. The closed-set classifier $\mathcal{C}$ is trained with a standard cross-entropy loss, $\mathcal{L}_{cls}$. The open-set classifiers $\mathcal{O}$ are trained using Hard Negative Classifier Sampling~(HNCS), which encourages each classifier $\mathcal{O}^k$ to learn a tight boundary between its corresponding class $k$ and the most confusing negative class. This loss, $\mathcal{L}_{ova}$, for a source sample $(x^s_i, y^s_i)$ is given by~\cite{saito2021ovanet}:
\begin{equation}
    \mathcal{L}_{ova}(x^s_i, y^s_i) = -\log p_o(y^s_i|x^s_i) - \min_{k \neq y^s_i} \log(1 - p_o(k|x^s_i)),
\end{equation}
where $p_o(k|x^s_i)$ is the score from the $k$-th open-set classifier, $\mathcal{O}^k$, indicating the probability that $x^s_i$ is an in-lier for class $k$. For adaptation, OVANet uses Open-set Entropy Minimization~(OEM) on target samples to encourage confident predictions from the open-set classifiers~\cite{saito2021ovanet}:
\begin{multline}
    \mathcal{L}_{oem}(x^t_j) = -\frac{1}{K} \sum_{k=1}^{K} [ p_o(k|x^t_j)\log p_o(k|x^t_j) + \\ (1-p_o(k|x^t_j))\log(1-p_o(k|x^t_j)) ].
\end{multline}

\noindent\textbf{MLNet Extensions.} MLNet~\cite{lu2024mlnet} enhances OVANet with three additional constraints.
\begin{enumerate}
    \item \textbf{Neighborhood Invariance Learning ($\mathcal{L}_{nil}$):} This constraint reduces intra-domain variations in the target feature space. For a target sample $x^t_j$, it encourages its feature representation to be similar to its neighbors $N_j$ found in a memory bank $\mathcal{M}$ and is given by~\cite{lu2024mlnet}:
    \begin{equation}
        \mathcal{L}_{nil}(x^t_j) = -\frac{1}{|N_j|} \sum_{k \in N_j} w_{jk} \log p_{jk},
    \end{equation}
    where $p_{jk}$ is the softmax-normalized similarity (i.e., softmax of a dot product) between the $l_2$-normalized features of $z^t_j = \mathcal{F}(x^t_j)$ and its neighbor $z^t_k = \mathcal{F}(x^t_k)$, and $w_{jk} = \frac{|N_j \cap N_k|}{|N_j \cup N_k|}$ is a confidence weight based on the Jaccard similarity of their respective neighborhoods. \Add{By pulling a sample's feature representation closer to its high-confidence neighbors, which are likely to share the same class, the feature cluster for that class becomes tighter, making classes more separable.}

    \item \textbf{Cross-domain Manifold Mixup ($\mathcal{L}_{cmm}$):} This constraint simulates ``unknown'' samples by creating a mixed feature $z^m_{i,j,\lambda} = \lambda z^s_i + (1 - \lambda) z^t_j$ through a linear combination of a source feature $z^s_i = \mathcal{F}(x^s_i)$ and a target feature $z^t_j = \mathcal{F}(x^t_j)$ with a weight $\lambda \sim Beta(\alpha, \alpha), \alpha = 2.0$. The \Add{act of mixing features across different domains is sufficient to generate a challenging out-of-distribution sample for the} open-set classifier related to the source class $y^s_i$\Add{, which} is then trained to reject this sample~\cite{lu2024mlnet}:
    \begin{equation}
        \mathcal{L}_{cmm}(x^s_i, y^s_i, x^t_j) = -\log(1 - p_o(y^s_i|z^m_{i,j,\lambda})).
    \end{equation}

    \item \textbf{Consistency Constraint ($\mathcal{L}_{cc}$):} This constraint counteracts the tendency of $\mathcal{L}_{cmm}$ to misclassify knowns as unknown. It identifies target samples \Rep{with high closed-set confidence but low open-set confidence and pushes the open-set score higher}{where the closed-set confidence, $p_c(k|x^t_j)$, is high, but the open-set score, $p_o(k|x^t_j)$, is low. The loss is defined as}~\cite{lu2024mlnet}:
    \begin{equation}
        \mathcal{L}_{cc}(x^t_j) = -\frac{1}{K} \sum_{k=1}^{K} p_c(k|x^t_j) \cdot p_o(k|x^t_j).
    \end{equation}
    \Add{This pushes the open-set score $p_o(k|x^t_j)$ to increase by a magnitude proportional to the closed-set confidence $p_c(k|x^t_j)$, thus correcting misidentified known samples.}

\end{enumerate}
The overall training loss of MLNet combines all of these components, including the  $\mathcal{L}_{cls}$ and $\mathcal{L}_{oem}$ constraints of OVANet.

\subsection{Enhanced MLNet}
The standard OEM constraint, $\mathcal{L}_{oem}$, averages the binary entropy over all $K$ open-set classifiers, $\mathcal{O}$, assuming that all source classes are equally relevant for any given target sample. This overlooks the fact that a sample's feature representation will be close to some decision boundaries but far from others, thereby diluting the learning signal by including irrelevant classifiers in the adaptation.

Motivated by this observation, we propose to enhance MLNet by using the predictions of the closed-set classifier $\mathcal{C}$ as dynamic, sample-specific weights for the open-set entropy terms. This allows the model to focus adaptation on the most relevant classes. Inspired by~\cite{yu2024selfcalibrated}, we reformulate the $\mathcal{L}_{oem}$ constraint for a target sample $x^t_j$ as:
\begin{multline}
    \mathcal{L}_{oem}(x^t_j) = -\frac{1}{K} \sum_{k=1}^{K} p_c(k|x^t_j) \cdot [ p_o(k|x^t_j)\log p_o(k|x^t_j) + \\ (1-p_o(k|x^t_j))\log(1-p_o(k|x^t_j)) ],
\end{multline}
\Add{where $p_c(k|x^t_j)$ is its prediction probability for class $k$ from the closed-set classifier $\mathcal{C}$.}
This formulation acts as \Rep{an attention mechanism}{a dynamic weighting strategy}. If the closed-set classifier $\mathcal{C}$ is confident that $x^t_j$ likely belongs to class $k$, its prediction $p_c(k|x^t_j)$ will be high. Consequently, the loss will be dominated by the term that encourages a confident in-lier/out-lier decision from the corresponding open-set classifier, $\mathcal{O}^k$. Entropy terms for other, less relevant classifiers will be down-weighted, preventing them from diluting the adaptation signal.

Following MLNet, the overall loss $\mathcal{L}_{all}$ is a weighted sum of all components~\cite{lu2024mlnet}:
\begin{equation}
\begin{split}
    \mathcal{L}_{all} = & \mathop{\mathbb{E}}_{(x^s_i, y^s_i) \sim D_s} [\mathcal{L}_{cls}(x^s_i, y^s_i) + \mathcal{L}_{ova}(x^s_i, y^s_i)] \\
    & + \mathop{\mathbb{E}}_{x^t_j \sim D_t} [\beta_1 \mathcal{L}_{nil}(x^t_j) + \eta \mathcal{L}_{cc}(x^t_j) + \gamma \mathcal{L}_{oem}(x^t_j)] \\
    & + \mathop{\mathbb{E}}_{(x^s_i,y^s_i)\sim D_s, x^t_j\sim D_t} [\beta_2 \mathcal{L}_{cmm}(x^s_i, y^s_i, x^t_j)],
\end{split}
\label{eq:final_loss}
\end{equation}
where $\beta_1, \beta_2, \eta,$ and $\gamma$ are hyperparameters that balance the contributions of the different loss components. For a fair comparison with MLNet, we adopt their settings for $\beta_1, \beta_2,$ and $\eta$, and set $\gamma$ following OVANet. 

\section{Experiments and Results}
\label{sec:setup}
This section presents our experiments and their results. We first detail the experimental setup and then report results for different UniDA settings on four well-known benchmarks.

\subsection{Datasets}
Our experiments are conducted on four widely-used domain adaptation benchmarks.
\begin{itemize}
    \item \textbf{Office-31}~\cite{saenko2010adapting} is a standard, medium-scale dataset containing 4,110 images across 31 categories, collected from three distinct domains: Amazon~(A), DSLR~(D), and Webcam~(W).
    \item \textbf{Office-Home}~\cite{venkateswara2017deep} is a more challenging dataset comprising 15,500 images across 65 categories. It features four domains with significant visual shifts: Art~(A), Clipart~(C), Product~(P), and Real-World~(R).
    \item \textbf{VisDA-2017}~\cite{peng2017visda} is a large-scale benchmark for simulation-to-real adaptation, consisting of 152,397 synthetic images and 55,388 real-world images distributed across 12 categories.
    \item \textbf{ImageCLEF}\footnote{\url{http://imageclef.org/2014/adaptation/} (As of \today)} is a dataset of 2,400 images across 12 common categories, sourced from four publicly available datasets: Bing~(B), Caltech-256~(C), ImageNet~(I), and PASCAL VOC-2012~(P). 
\end{itemize}

We follow the standard UniDA protocols established in prior work~\cite{lu2024mlnet, you2019universal} and evaluate on various category-shift scenarios. Due to space constraints, we focus our reporting on the OPDA and ODA settings. We omit CDA and PDA results, as our proposed modification to the OEM loss does not affect that specific scenario. For fair comparison, the class splits for each scenario are kept consistent with prior work~\cite{lu2024mlnet, you2019universal}. In the results section, each table specifies the class split using the format: $|L_s \cap L_t| / |L_s \setminus L_t| / |L_t \setminus L_s|$ (shared / source-private / target-private classes).

\subsection{Evaluation Metrics}
In UniDA, a model's performance must reflect a balance between correctly classifying known-class samples and accurately identifying unknown ones. A single accuracy metric can therefore be misleading. We adopt the standard metric for this task, the H-score ($\mathrm{HSC}$), which is the harmonic mean of the known-class and unknown-class accuracies~\cite{bucci2020effectiveness}:

\begin{equation}
    \mathrm{HSC} = \frac{2 \times \mathrm{OS*} \times \mathrm{UNK}}{\mathrm{OS*} + \mathrm{UNK}}.
\end{equation}
\noindent where $\mathrm{OS*}$ is the average per-class accuracy on the set of shared categories ($L_s \cap L_t$) and $\mathrm{UNK}$ is the accuracy on the unknown categories ($L_t \setminus L_s$).
This metric yields a high score only when both of its components are high, providing a balanced measure of performance on the UniDA task.

\subsection{Implementation Details}
\iffinal
For a fair comparison, our implementation\footnote{\Add{Our code is available at \url{https://github.com/jurandy-almeida/E-MLNet}}} is built directly upon the official MLNet framework\footnote{\url{https://github.com/YanzuoLu/MLNet} (Accessed: \today)}.
\fi
We use a ResNet-50\Del{~\cite{he2016deep}} model pre-trained on ImageNet as the backbone feature extractor $\mathcal{F}$, replacing its final fully-connected layer with new classification heads. The network is trained using an SGD optimizer with Nesterov momentum of 0.9 and a weight decay of $5 \times 10^{-4}$. We adopt an inverse learning rate decay schedule, with an initial learning rate of $1 \times 10^{-3}$ for the backbone and $1 \times 10^{-2}$ for the new layers. All experiments were run for 50 epochs with a batch size of 36.

The hyperparameters for the MLNet loss components were set according to the original study to ensure a direct comparison~\cite{lu2024mlnet}: $\beta_1=0.5$ for $\mathcal{L}_{nil}$, $\beta_2=0.1$ for $\mathcal{L}_{cmm}$, and $\eta=0.16$ for $\mathcal{L}_{cc}$ (doubled for the VisDA-2017 dataset). For the OEM loss hyperparameter, $\gamma$, we follow OVANet~\cite{saito2021ovanet} and set it to 0.1 across all experiments.

\Rep{To ensure reproducibility, all}{All} experiments were conducted on a single machine with \Del{the following configuration:} an AMD Ryzen Threadripper PRO 5975WX 32-core CPU, 256 GB of DDR4 memory, and four NVIDIA RTX 5000 Ada Generation GPUs. The system runs Ubuntu 22.04.4 LTS with Linux kernel 6.8.0 and the ext4 file system.

\begin{table*}[!htb]
    \caption{H-score (\%) comparison in the OPDA setting for Office-31, Office-Home, and VisDA.}
    \label{tab:opda_results}
    \centering
    \resizebox{\textwidth}{!}{
    \begin{tabular}{lccccccccccccccccccccc}
        \toprule
        \multirow{2}{*}{\textbf{Method}} & \multicolumn{7}{c}{\textbf{Office-31 (10/10/11)}} & \multicolumn{13}{c}{\textbf{Office-Home (10/5/50)}} & \textbf{VisDA} \\
        \cmidrule(lr){2-8} \cmidrule(lr){9-21}
        & A2D & A2W & D2A & D2W & W2A & W2D & Avg & A2C & A2P & A2R & C2A & C2P & C2R & P2A & P2C & P2R & R2A & R2C & R2P & Avg & \textbf{(6/3/3)} \\
        \midrule
        UAN~\cite{you2019universal} & 59.7 & 58.6 & 60.1 & 70.6 & 60.3 & 71.4 & 63.5 & 51.6 & 51.7 & 54.3 & 61.7 & 57.6 & 61.9 & 50.4 & 47.6 & 61.5 & 62.9 & 52.6 & 65.2 & 56.6 & 30.5 \\
        CMU~\cite{fu2020learning} & 68.1 & 67.3 & 71.4 & 79.3 & 72.2 & 80.4 & 73.1 & 56.0 & 56.6 & 59.2 & 67.0 & 64.3 & 67.8 & 54.7 & 51.1 & 66.4 & 68.2 & 57.9 & 69.7 & 61.6 & 34.6 \\
        DANCE~\cite{saito2020universal} & 79.6 & 75.8 & 82.9 & 90.9 & 77.6 & 87.1 & 82.3 & 61.0 & 60.4 & 64.9 & 65.7 & 58.8 & 61.8 & 73.1 & 61.2 & 66.6 & 67.7 & 62.4 & 63.7 & 63.9 & 42.8 \\
        DCC~\cite{li2021domain} & 88.5 & 78.5 & 70.2 & 79.3 & 75.9 & 88.6 & 80.2 & 58.0 & 54.1 & 58.0 & 74.6 & 70.6 & 77.5 & 64.3 & 73.6 & 74.9 & 81.0 & 75.1 & 80.4 & 70.2 & 43.0 \\
        GATE~\cite{chen2022geometric} & 87.7 & 81.6 & 84.2 & 94.8 & 83.4 & 94.1 & 87.6 & 63.8 & 75.9 & 81.4 & 74.0 & 72.1 & 79.8 & 74.7 & 70.3 & 82.7 & 79.1 & 71.5 & 81.7 & 75.6 & 56.4 \\
        TNT~\cite{chen2022evidential} & 85.7 & 80.4 & 83.8 & 92.0 & 79.1 & 91.2 & 85.4 & 61.9 & 74.6 & 80.2 & 73.5 & 71.4 & 79.6 & 74.2 & 69.5 & 82.7 & 77.3 & 70.1 & 81.2 & 74.7 & 55.3 \\
        UniOT~\cite{chang2022unified} & 87.0 & 88.5 & 88.4 & 98.8 & 87.6 & 96.6 & 91.2 & 67.3 & 80.5 & 86.0 & 73.5 & 77.3 & 84.3 & 75.5 & 63.3 & 86.0 & 77.8 & 65.4 & 81.9 & 76.6 & 57.3 \\
        CPR~\cite{hur2023learning} & 84.4 & 81.4 & 85.5 & 93.4 & 91.3 & 96.8 & 88.8 & 59.0 & 77.1 & 83.7 & 69.7 & 68.1 & 75.4 & 74.6 & 56.1 & 78.9 & 80.5 & 63.0 & 81.0 & 72.3 & 58.2 \\
        NCAL~\cite{su2023neighborhood} & 85.3 & 85.3 & 88.0 & 94.0 & 87.9 & 95.5 & 89.3 & 59.1 & 88.3 & 87.3 & 72.1 & 73.2 & 81.0 & 76.3 & 57.4 & 88.4 & 81.1 & 62.0 & 85.4 & 75.9 & 62.9 \\
        \midrule
        OVANet~\cite{saito2021ovanet} & 83.9 & 78.5 & 81.3 & 95.4 & 83.2 & 96.4 & 86.5 & 62.2 & 79.0 & 80.0 & 69.2 & 70.5 & 76.5 & 70.9 & 59.5 & 80.9 & 76.8 & 62.8 & 79.6 & 72.3 & 49.6 \\
        MLNet~\cite{lu2024mlnet} & \underline{90.4} & \textbf{93.7} & \textbf{89.7} & \underline{96.2} & \textbf{88.4} & \textbf{98.3} & \textbf{92.8} & \underline{68.2} & \textbf{83.8} & \textbf{85.0} & \underline{73.6} & \textbf{78.2} & \textbf{82.2} & \underline{75.2} & \underline{64.7} & \textbf{85.1} & \textbf{78.8} & \textbf{69.9} & \underline{83.9} & \textbf{77.4} & \underline{69.9} \\
        \textbf{E-MLNet} & \textbf{91.8} & \underline{91.8} & \underline{89.4} & \textbf{98.3} & \underline{88.2} & \underline{96.5} & \underline{92.6} & \textbf{68.6} & \underline{82.8} & \underline{84.2} & \textbf{73.8} & \textbf{78.2} & \textbf{82.2} &  \textbf{75.4} & \textbf{65.2} & \underline{84.4} & \underline{77.8} & \underline{69.6} & \textbf{84.0} & \underline{77.2} & \textbf{70.3} \\
        \bottomrule
    \end{tabular}
    }
\end{table*}

\begin{table*}[!htb]
    \caption{H-score (\%) comparison in the OPDA setting for ImageCLEF.}
    \label{tab:opda_results_imageclef}
    \centering
    \resizebox{0.62\textwidth}{!}{
    \begin{tabular}{lccccccccccccc}
        \toprule
        \multirow{2}{*}{\textbf{Method}} & \multicolumn{13}{c}{\textbf{ImageCLEF (6/3/3)}} \\
        \cmidrule(lr){2-14}
        & B2C & B2I & B2P & C2B & C2I & C2P & I2B & I2C & I2P & P2B & P2C & P2I & Avg \\
        \midrule
        OVANet~\cite{saito2021ovanet} & 68.9 & 70.4 & 63.4 & 62.0 & 81.3 & 74.3 & 59.5 & 73.4 & 66.8 & 54.4 & 83.1 & 66.9 & 68.7 \\        
        MLNet~\cite{lu2024mlnet} & \underline{96.7} & \underline{77.0} & \underline{64.6} & \underline{68.8} & \underline{87.7} & \underline{66.6} & \underline{65.9} & \underline{93.7} & \underline{80.1} & \underline{64.7} & \underline{89.8} & \underline{81.8} & \underline{78.1} \\
        \textbf{E-MLNet} & \textbf{97.3} & \textbf{79.5} & \textbf{66.1} & \textbf{70.6} & \textbf{89.0} & \textbf{69.3} & \textbf{67.1} & \textbf{94.3} & \textbf{81.4} & \textbf{64.9} & \textbf{90.2} & \textbf{88.3} & \textbf{79.8} \\
        \bottomrule
    \end{tabular}
    }
\end{table*}

\begin{table*}[!htb]
    \caption{H-score (\%) comparison in the ODA setting for Office-31, Office-Home, and VisDA.}
    \label{tab:oda_results}
    \centering
    \resizebox{\textwidth}{!}{
    \begin{tabular}{lccccccccccccccccccccc}
        \toprule
        \multirow{2}{*}{\textbf{Method}} & \multicolumn{7}{c}{\textbf{Office-31 (10/0/11)}} & \multicolumn{13}{c}{\textbf{Office-Home (25/0/40)}} & \textbf{VisDA} \\
        \cmidrule(lr){2-8} \cmidrule(lr){9-21}
        & A2D & A2W & D2A & D2W & W2A & W2D & Avg & A2C & A2P & A2R & C2A & C2P & C2R & P2A & P2C & P2R & R2A & R2C & R2P & Avg & \textbf{(6/0/6)} \\
        \midrule
        UAN~\cite{you2019universal} & 38.9 & 46.8 & 68.0 & 68.8 & 54.9 & 53.0 & 55.1 & 40.3 & 41.5 & 46.1 & 53.2 & 48.0 & 53.7 & 40.6 & 39.8 & 52.5 & 53.6 & 43.7 & 56.9 & 47.5 & 51.9 \\
        CMU~\cite{fu2020learning} & 52.6 & 55.7 & 76.5 & 75.9 & 65.8 & 64.7 & 65.2 & 45.1 & 48.3 & 51.7 & 58.9 & 55.4 & 61.2 & 46.5 & 43.8 & 58.0 & 58.6 & 50.1 & 61.8 & 53.3 & 54.2 \\
        DANCE~\cite{saito2020universal} & 84.9 & 78.8 & 79.1 & 78.8 & 68.3 & 88.9 & 79.8 & 61.9 & 61.3 & 63.7 & 64.2 & 58.6 & 62.6 & 67.4 & 61.0 & 65.5 & 65.9 & 61.3 & 64.2 & 63.1 & 67.5 \\
        DCC~\cite{li2021domain} & 58.3 & 54.8 & 67.2 & 89.4 & 85.3 & 80.9 & 72.7 & 56.1 & 67.5 & 66.7 & 49.6 & 66.5 & 64.0 & 55.8 & 53.0 & 70.5 & 61.6 & 57.2 & 71.9 & 61.7 & 59.6 \\
        GATE~\cite{chen2022geometric} & 88.4 & 86.5 & 84.2 & 95.0 & 86.1 & 96.7 & 89.5 & 63.8 & 70.5 & 75.8 & 66.4 & 67.9 & 71.7 & 67.3 & 61.5 & 76.0 & 70.4 & 61.8 & 75.1 & 69.0 & 70.8 \\
        TNT~\cite{chen2022evidential} & 85.8 & 82.3 & 80.7 & 91.2 & 81.5 & 96.2 & 86.3 & 63.4 & 67.9 & 74.9 & 65.7 & 67.1 & 68.3 & 64.5 & 58.1 & 73.2 & 67.8 & 61.9 & 74.5 & 67.3 & 71.6 \\
        NCAL~\cite{su2023neighborhood} & 84.0 & 93.4 & 93.4 & 85.4 & 89.0 & 87.2 & 88.7 & 64.2 & 74.1 & 80.5 & 68.1 & 72.5 & 77.0 & 66.9 & 58.1 & 79.1 & 74.6 & 63.5 & 79.6 & 71.5 & 69.1 \\        
        \midrule
        OVANet~\cite{saito2021ovanet} & 88.2 & 88.7 & 86.9 & 97.8 & 89.4 & 98.8 & 91.6 & 58.7 & 66.5 & 70.4 & 61.5 & 65.4 & 68.4 & 60.4 & 53.5 & 70.0 & 68.2 & 59.1 & 67.0 & 64.1 & 61.6 \\
        MLNet~\cite{lu2024mlnet} & \underline{93.0} & \underline{91.9} & \underline{86.9} & \underline{98.1} & \underline{87.5} & \textbf{99.5} & \underline{92.8} & \underline{61.3} & \textbf{69.9} & \textbf{74.4} & \textbf{63.1} & \textbf{68.2} & \textbf{70.4} & \textbf{62.0} & \textbf{59.9} & \textbf{72.4} & \textbf{69.1} & \textbf{62.6} & \textbf{71.1} & \textbf{67.0} & \underline{63.9} \\
        \textbf{E-MLNet} & \textbf{93.6} & \textbf{92.4} & \textbf{88.6} & \textbf{98.2} & \textbf{87.9} & \underline{98.6} & \textbf{93.2} & \textbf{61.6} & \underline{69.2} & \underline{73.1} & \underline{63.0} & \underline{67.6} & \underline{70.2} & \underline{61.7} & \underline{59.2} & \underline{71.9} & \underline{68.3} & \underline{62.1} & \underline{70.6} & \underline{66.6} & \textbf{66.6} \\
        \bottomrule
    \end{tabular}
    }
\end{table*}

\begin{table*}[!htb]
    \caption{H-score (\%) comparison in the ODA setting for ImageCLEF.}
    \label{tab:oda_results_imageclef}
    \centering
    \resizebox{0.62\textwidth}{!}{
    \begin{tabular}{lccccccccccccc}
        \toprule
        \multirow{2}{*}{\textbf{Method}} & \multicolumn{13}{c}{\textbf{ImageCLEF (6/0/6)}} \\
        \cmidrule(lr){2-14}
        & B2C & B2I & B2P & C2B & C2I & C2P & I2B & I2C & I2P & P2B & P2C & P2I & Avg \\
        \midrule
        OSNN~\cite{mendes2017nearest} & 74.9 & 69.1 & 63.5 & 54.9 & 60.4 & 57.3 & 49.8 & 57.0 & 54.2 & 52.2 & 67.3 & 64.1 & 60.4 \\
        STA~\cite{liu2019separate} & 66.5 & 71.2 & 59.8 & 65.2 & 77.2 & 65.7 & 57.9 & 68.4 & 68.2 & 51.0 & 63.2 & 65.1 & 65.0 \\
        OSBP~\cite{saito2018open} & 83.9 & 74.3 & 66.5 & 59.9 & 84.3 & 66.7 & 58.1 & 86.3 & 70.1 & 56.3 & 78.9 & 72.6 & 71.5 \\
        DAOD~\cite{fang2020open} & 80.7 & 84.3 & 76.3 & 49.1 & 83.6 & 76.7 & 55.7 & 81.2 & 76.9 & 51.3 & 80.5 & 83.9 & 73.3 \\
        ROS~\cite{bucci2020effectiveness} & 83.8 & 74.6 & 62.9 & 63.3 & 80.6 & 73.3 & 58.8 & 90.6 & 77.0 & 52.7 & 79.7 & 80.5 & 73.1 \\
        \midrule
        OVANet~\cite{saito2021ovanet} & 74.1 & 79.6 & 68.7 & 66.5 & \textbf{86.3} & \textbf{76.8} & \textbf{64.9} & 86.0 & 74.2 & 55.7 & 65.7 & 61.5 & 71.7 \\        
        MLNet~\cite{lu2024mlnet} & \underline{88.8} & \underline{78.6} & \underline{69.0} & \underline{59.6} & 82.4 & 63.8 & 60.4 & \underline{95.6} & \underline{80.3} & \underline{58.9} & \underline{89.1} & \underline{86.3} & \underline{76.1} \\
        \textbf{E-MLNet} & \textbf{89.1} & \textbf{80.4} & \textbf{70.1} & \textbf{61.6} & \underline{85.1} & \underline{65.4} & \underline{62.2} & \textbf{96.3} & \textbf{81.3} & \textbf{60.4} & \textbf{89.8} & \textbf{87.3} & \textbf{77.4} \\
        \bottomrule
    \end{tabular}
    }
\end{table*}

\subsection{Results on UniDA}

We first evaluate all methods in the OPDA setting, the most challenging UniDA scenario, which includes both source-private and target-private classes. The results are presented in Tables~\ref{tab:opda_results}~and~\ref{tab:opda_results_imageclef}. For the family of methods based on one-vs-all classifiers (OVANet, MLNET, and E-MLNet), the best results are in \textbf{bold} and the second-best ones are in \underline{underline}.

Our analysis first focuses on the family of one-vs-all methods: OVANet, MLNet, and our E-MLNet. Both MLNet and E-MLNet significantly outperform OVANet across all datasets, with average H-score gains of at least 6.1\% on Office-31, 4.9\% on Office-Home, 20.3\% on VisDA, and 9.4\% on ImageCLEF. 

The primary comparison is between E-MLNet and its direct predecessor, MLNet. While both methods achieve similar average H-scores, E-MLNet demonstrates superior robustness, outperforming MLNet in 22 of the 31 individual adaptation tasks. This highlights the benefit of using the weighted strategy: by dynamically focusing the adaptation on relevant class boundaries, E-MLNet achieves more consistent performance across a wider range of scenarios—a critical attribute for a truly ``universal'' method. Notably, E-MLNet obtains the highest average H-score on the challenging VisDA and ImageCLEF benchmarks, underscoring the effectiveness of our approach. 
\Add{We hypothesize that the few cases where MLNet performs better occur in tasks with severe domain shifts where the closed-set classifier's initial predictions are less reliable. In such cases, uniform weighting of the standard OEM loss is a `safer' albeit less efficient strategy.}

Next, we analyze the ODA setting, where the target domain contains unknown classes but all source classes are shared. The results are detailed in Tables~\ref{tab:oda_results}~and~\ref{tab:oda_results_imageclef}.

E-MLNet and MLNet again surpass OVANet on all datasets, confirming the strength of the MLNet framework. Crucially, even without source-private classes, E-MLNet maintains an edge in consistency, achieving a higher H-score than MLNet in 19 out of 31 tasks. This result confirms that our focused adaptation strategy is beneficial even when the open-set recognition task is simpler, leading to more robust decision boundaries. When compared to all methods, the OVA-based family is highly competitive, though methods like GATE and NCAL also show strong results on the Office-Home dataset.

\section{Conclusion}

In this paper, we proposed the Enhanced Mutual Learning Network (E-MLNet), a method that refines the adaptation process of MLNet by integrating a dynamic, sample-specific weighting strategy for OEM.
Inspired by~\cite{yu2024selfcalibrated}, E-MLNet employs the predictions from the closed-set classifier to modulate the entropy contribution of each open-set classifier. 
This simple yet effective modification sharpens the learned feature representations, improving the model's ability to distinguish between known shared classes and unknown private classes in the target domain.
Our extensive experiments on four challenging benchmarks demonstrated the tangible benefits of this approach. E-MLNet showed superior robustness and consistency over its strong baseline, outperforming MLNet in the majority of individual tasks (22 of 31 in the OPDA setting and 19 of 31 in the ODA setting) and achieved the highest average H-scores on VisDA and ImageCLEF.

Despite its good performance, E-MLNet has limitations. Its effectiveness is highly dependent on the accuracy of the closed-set classifier. In cases of severe domain shift where the closed-set classifier's initial predictions are unreliable, our method could potentially focus adaptation on incorrect class boundaries, leading to error propagation. This dependency suggests that performance could be hindered in the very early stages of training before the model has stabilized.

For future work, this \Add{dynamic} weighting strategy could be explored further. Mitigating the identified limitations by introducing a confidence threshold for applying the weights, or developing more robust guidance mechanisms that are less sensitive to initial classification errors, presents a promising research direction. In addition, \Rep{this idea could be extended to other related open-set recognition problems beyond domain adaptation.}{strategies like Outlier Exposure~\cite{hendrycks2018deep} could be explored as an alternative to OEM.}

\iffinal
\section*{Acknowledgment}

This research was supported by São Paulo Research Foundation - FAPESP (grants 2023/17577-0, 2024/04500-2, and 2024/22985-3) and National Council for Scientific and Technological Development - CNPq (grants 315220/2023-6, 420442/2023-5, and 444982/2024-8).

\fi

\bibliographystyle{IEEEtran}



\end{document}